\documentclass[11pt]{article}
\usepackage{acl2015}
\usepackage{times}
\usepackage{url}
\usepackage{latexsym}
\usepackage{xcolor}
\usepackage{balance}
\usepackage{graphicx}
\usepackage{epstopdf}
\usepackage{amsmath}
\usepackage{color}
\usepackage{amsfonts}
\usepackage{amssymb}
\usepackage{booktabs}
\usepackage{txfonts}
\usepackage{enumitem}

\newcommand{\denselist}{\setlength{\itemsep}{1pt} \setlength{\parskip}{0pt} \setlength{\parsep}{0pt}}
\newcommand{\bitem}{\begin{itemize}[noitemsep,topsep=2pt]\denselist}
\newcommand{\eitem}{\end{itemize}}

\DeclareMathOperator*{\argmax}{argmax}

\title{Structured Training for Neural Network Transition-Based Parsing}

\author{
  David Weiss \quad Chris Alberti \quad Michael Collins \quad Slav Petrov\\
  Google Inc\\
  New York, NY\\
  {\tt \{djweiss,chrisalberti,mjcollins,slav\}@google.com}
}

\date{}

\begin{document}
\maketitle


\begin{abstract}
  We present structured perceptron training for neural network
  transition-based dependency parsing. We learn the neural network
  representation using a gold corpus augmented by a large number of
  automatically parsed sentences. Given this fixed network
  representation, we learn a final layer using the structured
  perceptron with beam-search decoding. On the Penn Treebank, our
  parser reaches 94.26\% unlabeled and 92.41\% labeled attachment
  accuracy, which to our knowledge is the best accuracy on
  Stanford Dependencies to date.  We also provide in-depth ablative
  analysis to determine which aspects of our model provide the largest
  gains in accuracy.
\end{abstract}

\graphicspath{{./figs/}}
\newcommand{\bX}{\mathbf{X}}
\newcommand{\bE}{\mathbf{E}}
\newcommand{\bb}{\mathbf{b}}
\newcommand{\bH}{\mathbf{H}}
\newcommand{\bW}{\mathbf{W}}
\newcommand{\bh}{\mathbf{h}}
\newcommand{\mwords}{\mathrm{word}}
\newcommand{\mtags}{\mathrm{tag}}
\newcommand{\mlabels}{\mathrm{label}}
\newcommand{\todo}[1]{{\bf \color{red}{TODO: #1}}}
\newcommand{\eat}[1]{\ignorespaces}
\newcommand{\commentout}[1]{}
\newcommand\T{\rule{0pt}{4ex}}  




\section{Introduction}

Syntactic analysis is a central problem in language understanding that has received a tremendous amount of attention.
Lately, dependency parsing has emerged as a popular approach to this problem due to the availability of dependency treebanks in many languages \cite{Buchholz06,nilsson2007conll,mcdonald-EtAl:2013:ACL} and the efficiency of dependency parsers.

Transition-based parsers \cite{Nivre:2008:CL} have been shown to
provide a good balance between efficiency and accuracy.
In transition-based parsing, sentences are processed in a linear left to right pass; at each position, the parser needs to choose from a set of possible actions defined by the transition strategy.
In greedy models, a classifier is used to independently decide which transition to take based on local features of the current parse configuration.
This classifier typically uses hand-engineered features and is trained on individual transitions extracted from the gold transition sequence.
While extremely fast, these greedy models typically suffer from search errors due to the inability to recover from incorrect decisions.
\newcite{zhang2008tale} showed that a beam-search decoding algorithm utilizing the structured perceptron training algorithm can greatly improve accuracy.
Nonetheless, significant manual feature engineering was required before transition-based systems provided competitive accuracy with graph-based parsers \cite{zhang-nivre:2011:ACL}, and only by incorporating graph-based scoring functions were \newcite{bohnet2012best} able to exceed the accuracy of graph-based approaches.

In contrast to these carefully hand-tuned approaches, \newcite{chen-manning:2014:EMNLP}
recently presented a neural network version of a greedy transition-based parser.
In their model, a feed-forward neural network with a hidden layer is used to make the transition decisions.
The hidden layer has the power to learn arbitrary combinations of the atomic inputs, thereby eliminating the need for hand-engineered features.
Furthermore, because the neural network uses a distributed representation, it is able to model lexical, part-of-speech (POS) tag, and arc label similarities in a continuous space.
However, although their model outperforms its greedy hand-engineered counterparts, it is not competitive with state-of-the-art dependency parsers that are trained for structured search.

In this work, we combine the representational power of neural networks with the
superior search enabled by structured training and inference, making our parser
one of the most accurate dependency parsers to date.  Training and testing on
the Penn Treebank \cite{marcus:1993:CL}, our transition-based parser achieves
93.99\% unlabeled (UAS) / 92.05\% labeled (LAS) attachment accuracy,
outperforming the 93.22\% UAS / 91.02\% LAS of \newcite{zhang-mcdonald:2014:ACL}
and 93.27 UAS / 91.19 LAS of \newcite{bohnet2012best}.  In addition, by
incorporating unlabeled data into training, we further improve the accuracy of
our model to 94.26\% UAS / 92.41\% LAS (93.46\% UAS / 91.49\% LAS for our greedy
model).

In our approach we start with the basic structure of \newcite{chen-manning:2014:EMNLP}, but with a deeper architecture  and improvements to the optimization procedure. These modifications (Section \ref{sec:model}) increase the performance of the greedy model by as much as 1\%.
As in prior work, we train the neural network to model the probability of individual parse actions. However, we do not use these probabilities directly for prediction.
Instead, we use the activations from all layers of the neural network as the representation in a structured perceptron model that is trained with beam search and early updates (Section \ref{sec:training}).
On the Penn Treebank, this structured learning approach significantly improves parsing accuracy by 0.8\%.

An additional contribution of this work is an effective way to leverage unlabeled data.
Neural networks are known to perform very well in the presence of large amounts of training data; however, obtaining more expert-annotated parse trees is very expensive.
To this end, we generate large quantities of high-confidence parse trees by parsing unlabeled data with two different 
parsers and selecting only the sentences for which the two parsers produced the same trees (Section \ref{sec:uptraining}).
This approach is known as ``tri-training'' \cite{li-etAl:2014:ACL} 
and we show that it benefits our neural network parser significantly more than other approaches.
By adding 10 million automatically parsed tokens to the training data, we improve the accuracy of our parsers by almost $\sim$1.0\% on web domain data.

We provide an extensive exploration of our model in Section
\ref{sec:discussion} through ablative analysis and other retrospective
experiments. One of the goals of this work is to provide guidance for
future refinements and improvements on the architecture and modeling
choices we introduce in this paper.

Finally, we also note that neural network representations have a long history in
syntactic parsing \cite{henderson:2004:ACL,titov2007fast,titov2010latent};
however, like \newcite{chen-manning:2014:EMNLP}, our network avoids any
recurrent structure so as to keep inference fast and efficient and to allow the
use of simple backpropagation to compute gradients.  Our work is also not the
first to apply structured training to neural networks (see
e.g. \newcite{conditional_neural_fields} and \newcite{neural_crf} for
Conditional Random Field (CRF) training of neural networks). Our paper extends
this line of work to the setting of inexact search with beam decoding for
dependency parsing; \newcite{zhou2015} concurrently explored a similar
approach using a structured probabilistic ranking objective.  \newcite{dyer2015}
concurrently developed the Stack Long Short-Term Memory (S-LSTM) architecture,
which does incorporate recurrent architecture and look-ahead, and which yields
comparable accuracy on the Penn Treebank to our greedy model.


\begin{figure}[t]
  \centering
  \includegraphics[scale=0.6]{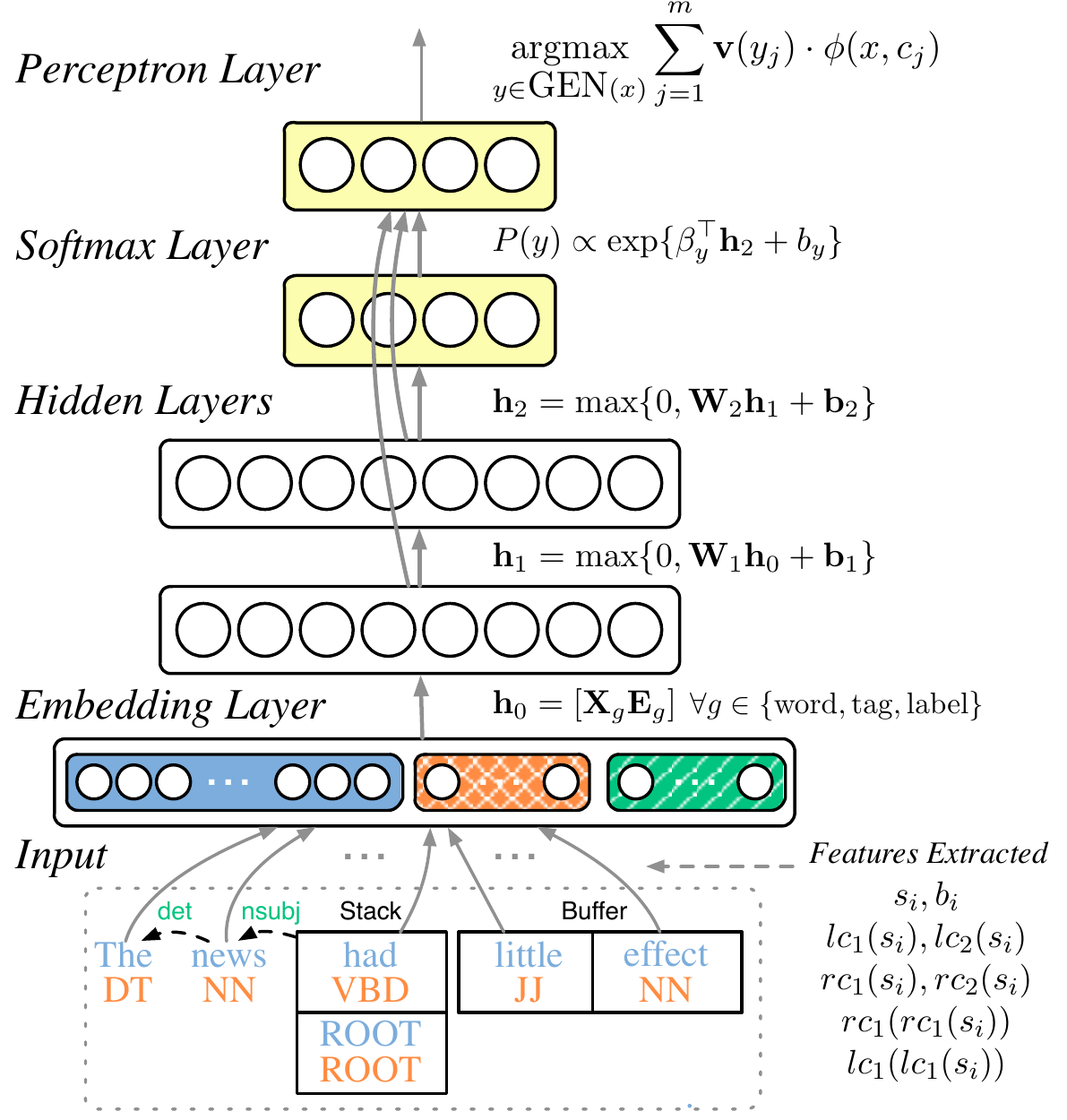}
  \caption{Schematic overview of our neural network model.
    Atomic features are extracted from the $i$'th elements
    on the stack ($s_i$) and the buffer ($b_i$); $lc_i$ indicates the $i$'th leftmost child
    and $rc_i$ the $i$'th rightmost child. We use the top two elements
    on the stack for the arc features and the top four tokens on stack and buffer for words, tags and arc labels.}
  \label{fig:schematic}
\end{figure}

\section{Neural Network Model}
\label{sec:model}

In this section, we describe the architecture of our model, which is
summarized in Figure \ref{fig:schematic}. Note that we separate the
embedding processing to a distinct ``embedding layer'' for clarity of
presentation.  Our model is based upon that of
\newcite{chen-manning:2014:EMNLP} and we discuss the differences
between our model and theirs in detail at the end of this section. We
use the {\em arc-standard} \cite{nivre2004incrementality} transition system.

\subsection{Input layer}

Given a parse configuration $c$ (consisting of a stack $s$ and a
buffer $b$), we extract a rich set of discrete features which we feed
into the neural network. Following \newcite{chen-manning:2014:EMNLP},
we group these features by their input source: words, POS tags, and
arc labels. The features extracted for each group are represented as a
sparse $F \times V$ matrix $\bX$, where $V$ is the size of the
vocabulary of the feature group and $F$ is the number of features. The
value of element $X_{fv}$ is $1$ if the $f$'th feature takes on value
$v$. We produce three input matrices: $\bX_{\mwords}$ for words
features, $\bX_{\mtags}$ for POS tag features, and $\bX_{\mlabels}$
for arc labels, with $F_{\mwords} = F_{\mtags} = 20$ and
$F_{\mlabels} = 12$ (Figure \ref{fig:schematic}).

For all feature groups, we add additional special values for ``ROOT''
(indicating the POS or word of the root token), ``NULL'' (indicating no valid
feature value could be computed) or ``UNK'' (indicating an out-of-vocabulary
item).

\subsection{Embedding layer}

The first learned layer $h_0$ in the network transforms the sparse,
discrete features $\bX$ into a dense, continuous embedded
representation. For each feature group $\bX_g$, we learn a
$V_g \times D_g$ embedding matrix $\bE_g$ that applies the
conversion:\begin{equation}
  \label{eq:embedding-layer}
  \bh_{0} = [\bX_{g}\bE_{g} \mid g \in \{\mwords,\mtags,\mlabels\}],
\end{equation}
where we apply the computation separately for each group $g$ and
concatenate the results. Thus, the embedding layer has
$E = \sum_g F_gD_g$ outputs, which we reshape to a vector $\bh_0$. 
We can choose the embedding dimensionality $D$
for each group freely. Since POS tags and arc labels have much smaller
vocabularies, we show in our experiments (Section \ref{sec:network_structure})
that we can use smaller $D_{\mtags}$ and $D_{\mlabels}$, without a
loss in accuracy.

\subsection{Hidden layers}

We experimented with one and two hidden layers composed of $M$
rectified linear (Relu) units \cite{relu}. Each unit in the hidden
layers is fully connected to the previous layer:
\begin{align}
  \label{eq:hidden-layer}
  \bh_{i} & = \max\{0, \bW_i\bh_{i-1} + \bb_i\},
\end{align}
where $\bW_1$ is a $M_1 \times E$ weight matrix for the first hidden layer and
$\bW_i$ are $M_i \times M_{i-1}$ matrices for all subsequent layers. The weights
$\bb_i$ are bias terms. Relu layers have been well studied in the neural network
literature and have been shown to work well for a wide domain of problems
\cite{krizhevsky2012,zeiler2013}. Through most of development, we kept
$M_i = 200$, but we found that significantly increasing the number of hidden
units improved our results for the final comparison.

\subsection{Relationship to \newcite{chen-manning:2014:EMNLP}\hspace*{-1cm}}

Our model is clearly inspired by and based on the work of
\newcite{chen-manning:2014:EMNLP}. There are a few structural
differences: (1) we allow for much smaller embeddings of POS tags and
labels, (2) we use Relu units in our hidden layers, and (3) we use a
deeper model with two hidden layers.
Somewhat to our surprise, we found these changes combined with an
SGD training scheme (Section \ref{sec:backprop}) during
the ``pre-training'' phase of the model to lead to an almost 1\%
accuracy gain over \newcite{chen-manning:2014:EMNLP}.
This trend held despite carefully tuning hyperparameters for
each method of training and structure combination.

Our main contribution from an algorithmic perspective is our training procedure: as
described in the next section, we use the structured perceptron for
learning the final layer of our model. 
We thus present a novel way to leverage a neural network representation
in a structured prediction setting.

\section{Semi-Supervised Structured Learning}
\label{sec:training}

In this work, we investigate a semi-supervised structured learning scheme that
yields substantial improvements in accuracy over the baseline neural network
model. There are two complementary contributions of our approach:
(1) incorporating structured learning of the model
and (2) utilizing unlabeled data. In both cases, we use the neural network to model the
probability of each parsing action $y$ as a soft-max function taking
the final hidden layer as its input:
\begin{equation}
  \label{eq:softmax}
  P(y) \propto \exp \{ \beta_y^\top \bh_i + b_y\},
\end{equation}
where $\beta_y$ is a $M_i$ dimensional vector of weights for class $y$
and $i$ is the index of the final hidden layer of the network. At a
high level our approach can be summarized as follows:
\begin{itemize}
\item First, we pre-train the network's hidden representations by
  learning probabilities of parsing actions.  Fixing the hidden
  representations, we learn an additional final output layer using the
  structured perceptron that uses the output of the network's hidden
  layers. In practice this improves accuracy by $\sim$0.6\% absolute.
\item Next, we show that we can supplement the gold data with a large
  corpus of high quality automatic parses. We show that incorporating
  unlabeled data in this way improves accuracy by as much as 1\%
  absolute.
\end{itemize}

\subsection{Backpropagation Pretraining}
\label{sec:backprop}

To learn the hidden representations, we use mini-batched averaged stochastic
gradient descent (ASGD) \cite{asgd} with momentum \cite{momentum} to learn the
parameters $\Theta$ of the network, where
$\Theta = \{\bE_g, \bW_i, \bb_i, \beta_y \mid \forall g,i,y\}$. We use
back-propagation to minimize the multinomial logistic loss:
\begin{equation}
  \label{eq:loss}
  L(\Theta) = -\sum_j \log  P(y_j \mid c_j, \Theta) + \lambda  \sum_i ||\bW_i||_2^2,
\end{equation}
where $\lambda$ is a regularization hyper-parameter over the hidden
layer parameters (we use $\lambda = 10^{-4}$ in all experiments) and
$j$ sums over all decisions and configurations $\{y_j,c_j\}$ extracted
from gold parse trees in the dataset.

The specific update rule we apply at iteration $t$ is as follows:
\begin{align}
  g_t &= \mu g_{t-1} - \Delta L(\Theta_t), \\
  \Theta_{t+1} & = \Theta_t + \eta_t g_t,
\end{align}
where the descent direction $g_t$ is computed by a weighted
combination of the previous direction $g_{t-1}$and the current
gradient $\Delta L(\Theta_t)$. The parameter $\mu \in [0,1)$ is the
momentum parameter while $\eta_t$ is the traditional learning
rate. In addition, since we did not tune the regularization parameter
$\lambda$, we apply a simple exponential step-wise decay to $\eta_t$;
for every $\gamma$ rounds of updates, we multiply
$\eta_t = 0.96\eta_{t-1}$.

The final component of the update is parameter averaging: we maintain
averaged parameters
$\bar{\Theta}_t = \alpha_t \bar{\Theta}_{t-1} + (1-\alpha_t)\Theta_t$,
where $\alpha_t$ is an averaging weight that increases from 0.1
to $0.9999$ with $1/t$. Combined with averaging, careful tuning of the three
hyperparameters $\mu$, $\eta_0$, and $\gamma$ using held-out data was
crucial in our experiments.

\subsection{Structured Perceptron Training}
\label{sec:perceptron}

\newcommand{\gen}{\hbox{GEN}}
\newcommand{\reals}{\mathbb{R}}

Given the hidden representations, we now describe how the
perceptron can be trained to utilize
these representations.  The perceptron algorithm with early updates
\cite{collins-roark:2004:ACL} requires a feature-vector definition
$\phi$ that maps a sentence $x$ together with a configuration $c$ to a
feature vector $\phi(x, c) \in \reals^d$. There is a one-to-one
mapping between configurations $c$ and decision sequences
$y_1 \ldots y_{j-1}$ for any integer $j \geq 1$: we will use $c$ and
$y_1 \ldots y_{j-1}$ interchangeably.

For a sentence $x$, define $\gen(x)$ to be the set of parse trees for
$x$. Each $y \in \gen(x)$ is a sequence of decisions $y_1 \ldots y_m$
for some integer $m$.
We use ${\cal Y}$ to denote the set of possible decisions in the
parsing model.  For each decision $y \in {\cal Y}$ we assume a
parameter vector $\mathbf{v}(y) \in \reals^d$. These parameters will be trained
using the perceptron.

In decoding with the perceptron-trained model, we will use beam search
to attempt to find:
\[
\argmax_{y \in \gen(x)} \sum_{j=1}^m \mathbf{v}(y_j) \cdot \phi(x, y_1 \ldots y_{j-1}).
\]
Thus each decision $y_j$ receives a score:
\[
\mathbf{v}(y_j) \cdot \phi(x, y_1 \ldots y_{j-1}).
\]

In the perceptron with early updates, the parameters $\mathbf{v}(y)$ are trained
as follows. On each training example, we run beam search until the
gold-standard parse tree falls out of the beam.\footnote{If the gold
  parse tree stays within the beam until the end of the
  sentence, conventional perceptron updates are used.}
Define $j$ to be the length of the beam at this point.
A structured perceptron update is performed using the gold-standard
decisions $y_1 \ldots y_j$ as the target, and the highest scoring
(incorrect) member of the beam as the negative example.

A key idea in this paper is to use the neural network to define the
representation $\phi(x, c)$. Given the sentence $x$ and the
configuration $c$, assuming two hidden layers, the neural network
defines values for $\bh_1$, $\bh_2$, and $P(y)$ for each decision
$y$. We experimented with various definitions of $\phi$ (Section
\ref{sec:perceptron_results}) and found that
$\phi(x,c) = [ \bh_1~\bh_2~P(y)]$ (the concatenation of the outputs
from both hidden layers, as well as the probabilities for all
decisions $y$ possible in the current configuration) had the best
accuracy on development data.

Note that it is possible to continue to use backpropagation to learn the
representation $\phi(x,c)$ during perceptron training; however, we found using
ASGD to pre-train the representation always led to faster, more accurate results
in preliminary experiments, and we left further investigation for future work.

\subsection{Incorporating Unlabeled Data}
\label{sec:uptraining}

Given the high capacity, non-linear nature of the deep network we
hypothesize that our model can be significantly improved by
incorporating more data.  One way to use unlabeled data is
through unsupervised methods such as word clusters
\cite{koo-etAl:2008:ACL}; we follow \newcite{chen-manning:2014:EMNLP}
and use pretrained word embeddings to initialize our model.  The word
embeddings capture similar distributional information as word clusters
and give consistent improvements by providing a
good initialization and information about words not seen in the treebank data.

However, obtaining more training data is even more important than
a good initialization. One potential way to obtain additional training data
is by parsing unlabeled data with previously
trained models.  \newcite{mcclosky-etAl:2006:NAACL} and
\newcite{huang-harper:2009:EMNLP} showed that iteratively
re-training a single model (``self-training'') can be used to improve parsers in certain settings;
\newcite{petrov-EtAl:2010:EMNLP} built on this work and showed that a
slow and accurate parser can be used to ``up-train'' a faster but less
accurate parser.

In this work, we adopt the ``tri-training'' approach of
\newcite{li-etAl:2014:ACL}: Two parsers are used to process the unlabeled corpus
and only sentences for which both parsers produced the same parse tree are added
to the training data.  The intuition behind this idea is that the chance of the
parse being correct is much higher when the two parsers agree: there is only one
way to be correct, while there are many possible incorrect parses. Of course,
this reasoning holds only as long as the parsers suffer from different biases.

We show that tri-training is far more effective than
vanilla up-training for our neural network model.  We use same setup
as \newcite{li-etAl:2014:ACL}, intersecting the output of the
BerkeleyParser \cite{petrov-EtAl:2006:ACL}, and a reimplementation of ZPar
\cite{zhang-nivre:2011:ACL} as our baseline parsers.
The two parsers agree only 36\% of the time on the tune set,
but their accuracy on those sentences is 97.26\% UAS, approaching
the inter annotator agreement rate.
These sentences are of course easier to parse,
having an average length of 15 words,
compared to 24 words for the tune set overall.
However, because we only use these sentences to extract individual transition decisions,
the shorter length does not seem to hurt their utility.
We generate $10^7$ tokens worth of new parses and use this data
in the backpropagation stage of training.


\section{Experiments}
\label{sec:final_experiments}

In this section we present our experimental setup and the main results
of our work.

\subsection{Experimental Setup}

We conduct our experiments on two English language benchmarks: (1) the standard
Wall Street Journal (WSJ) part of the Penn Treebank \cite{marcus:1993:CL} and
(2) a more comprehensive union of publicly available treebanks spanning multiple
domains. For the WSJ experiments, we follow standard practice and use sections 2-21 for
training, section 22 for development and section 23 as the final test set.
Since there are many hyperparameters in our models, we additionally use section
24 for tuning.  We convert the constituency trees to Stanford style dependencies
\cite{stanford_dependencies} using version 3.3.0 of the converter.  We use a
CRF-based POS tagger to generate 5-fold jack-knifed POS tags on the training set
and predicted tags on the dev, test and tune sets; our tagger gets comparable
accuracy to the Stanford POS tagger \cite{ToutanovaKMS03} with 97.44\% on the
test set.  We report unlabeled attachment score (UAS) and labeled attachment
score (LAS) excluding punctuation on predicted POS tags, as is standard for
English.

For the second set of experiments, we follow the same procedure as
above, but with a more diverse dataset for training and
evaluation. Following \newcite{vinyals2015}, we use (in addition to
the WSJ), the OntoNotes corpus version 5 \cite{hovy-EtAl:2006:NAACL},
the English Web Treebank \cite{petrov-mcdonald:2012:SANCL}, and the
updated and corrected Question Treebank \cite{judge-etAl:2006:ACL}. We
train on the union of each corpora's training set and test on each
domain separately. We refer to this setup as the ``Treebank Union''
setup.

In our semi-supervised experiments, we use the corpus from
\newcite{1-billion-lm} as our source of unlabeled data.  We process it with the
BerkeleyParser \cite{petrov-EtAl:2006:ACL}, a latent variable constituency
parser, and a reimplementation of ZPar \cite{zhang-nivre:2011:ACL}, a
transition-based parser with beam search.  Both parsers are included as
baselines in our evaluation.  We select the first $10^7$ tokens for which the
two parsers agree as additional training data. For our tri-training experiments,
we re-train the POS tagger using the POS tags assigned on the unlabeled data
from the Berkeley constituency parser. This increases POS accuracy slightly to
97.57\% on the WSJ.

\begin{table}
\hspace*{-0.6em}
\scalebox{0.9}{
\setlength{\tabcolsep}{4pt}
\begin{tabular}{lccc}
\toprule
{Method} & UAS & LAS & Beam \\
\midrule
{\em Graph-based} & & & \\
\quad \newcite{bohnet:2010:COLING} & 92.88 & 90.71 & n/a \\
\quad \newcite{martins-etAl:2013:ACL} & 92.89 & 90.55 & n/a \\
\quad \newcite{zhang-mcdonald:2014:ACL} & 93.22 & 91.02 & n/a\\
\midrule
{\em Transition-based} & & & \\
  \quad $^\star$\newcite{zhang-nivre:2011:ACL} & 93.00 & 90.95 & 32  \\
  \quad \newcite{bohnet2012best} & 93.27 & 91.19 & 40 \\
  \quad \newcite{chen-manning:2014:EMNLP} & 91.80 & 89.60 & 1 \\
  \quad S-LSTM \cite{dyer2015} & 93.20 & 90.90 & 1  \\
  \quad Our Greedy & 93.19 & 91.18 & 1  \\
  \quad Our Perceptron & {\bf 93.99} & {\bf 92.05} & 8 \\
\midrule
\midrule
 {\em Tri-training} & & & \\
\quad  $^\star$\newcite{zhang-nivre:2011:ACL} & 92.92 & 90.88 & 32 \\
\quad  Our Greedy & 93.46 & 91.49 & 1 \\
\quad  Our Perceptron  & {\bf 94.26} & {\bf 92.41} & 8 \\

\bottomrule
\end{tabular}}
\caption{\label{tab:english_final}
  Final WSJ test set results. We compare our system to state-of-the-art graph-based and transition-based dependency parsers.
  $^\star$ denotes our own re-implementation of the system so we could compare tri-training on a competitive baseline.
  All methods except \newcite{chen-manning:2014:EMNLP} and \newcite{dyer2015} were run using predicted tags from our POS tagger.
  For reference, the accuracy of the Berkeley constituency parser (after conversion) is 93.61\% UAS / 91.51\% LAS.}
\end{table}

\subsection{Model Initialization \& Hyperparameters}

In all cases, we initialized $\bW_i$ and $\beta$ randomly using a
Gaussian distribution with variance $10^{-4}$. We used fixed
initialization with $b_i = 0.2$, to ensure that most Relu units are
activated during the initial rounds of training. We did not
systematically compare this random scheme to others, but we found that
it was sufficient for our purposes.

For the word embedding matrix $\bE_{\mwords}$, we initialized the
parameters using pretrained word embeddings. We used the publicly
available {\tt word2vec}\footnote{http://code.google.com/p/word2vec/}
tool \cite{mikolov2013efficient} to learn CBOW embeddings following
the sample configuration provided with the tool. For
words not appearing in the unsupervised data and the special ``NULL''
etc. tokens, we used random initialization.  In preliminary
experiments we found no difference between training the word
embeddings on 1 billion or 10 billion tokens. We therefore trained the
word embeddings on the same corpus we used for tri-training \cite{1-billion-lm}.

We set $D_{\mwords} = 64$ and $D_{\mtags} = D_{\mlabels} = 32$ for embedding
dimensions and $M_1 = M_2 = 2048$ hidden units in our final experiments. For the
perceptron layer, we used $\phi(x,c) = [\bh_1~\bh_2~P(y)]$ (concatenation of all
intermediate layers). All hyperparameters (including structure) were tuned using
Section 24 of the WSJ only. When not tri-training, we used hyperparameters of
$\gamma = 0.2$, $\eta_0 = 0.05$, $\mu = 0.9$, early stopping after roughly 16
hours of training time. With the tri-training data, we decreased
$\eta_0 = 0.05$, increased $\gamma = 0.5$, and decreased the size of the network
to $M_1 = 1024$, $M_2 = 256$ for run-time efficiency, and trained the network
for approximately 4 days. For the Treebank Union setup, we set
$M_1 = M_2 = 1024$ for the standard training set and for the tri-training setup.

\begin{table}
  \centering
  \hspace*{-0.6em}
  \scalebox{0.9}{
    \setlength{\tabcolsep}{4pt}
    \begin{tabular}{lccc}
      \toprule
      {Method} & News & Web & QTB \\
      \midrule
      {\em Graph-based} & & & \\
      \quad \newcite{bohnet:2010:COLING} & 91.38 & 85.22 & 91.49 \\
      \quad \newcite{martins-etAl:2013:ACL} & 91.13 & 85.04 & 91.54 \\
      \quad \newcite{zhang-mcdonald:2014:ACL} & 91.48 & 85.59 & 90.69 \\
      \midrule
      {\em Transition-based} & & & \\
      \quad $^\star$\newcite{zhang-nivre:2011:ACL} & 91.15 & 85.24 & {\bf 92.46}  \\
      \quad \newcite{bohnet2012best} & 91.69 & 85.33 & 92.21 \\
      \quad Our Greedy & 91.21 & 85.41 & 90.61  \\
      \quad Our Perceptron ($B$=16) & {\bf 92.25} & {\bf 86.44} & 92.06 \\
      \midrule
      \midrule
      {\em Tri-training} & & & \\
      \quad  $^\star$\newcite{zhang-nivre:2011:ACL} & 91.46 & 85.51 & 91.36 \\
      \quad  Our Greedy & 91.82 & 86.37 & 90.58 \\
      \quad  Our Perceptron ($B$=16) & {\bf 92.62} & {\bf 87.00} & {\bf 93.05} \\
      \bottomrule
    \end{tabular}}
  \caption{\label{tab:treebank_union}
    Final Treebank Union test set results. We report LAS only for brevity; see Appendix for full results. For these tri-training results, we sampled sentences to ensure the distribution of sentence lengths matched the distribution in the training set, which we found marginally improved the ZPar tri-training performance.
    For reference, the accuracy of the Berkeley constituency parser (after conversion) is 91.66\% WSJ, 85.93\% Web, and 93.45\% QTB.}
\vspace{-0.8em}
\end{table}

\subsection{Results}

Table \ref{tab:english_final} shows our final results on the WSJ test
set, and Table \ref{tab:treebank_union} shows the cross-domain results
from the Treebank Union.  We compare to the best dependency parsers in
the literature. For \cite{chen-manning:2014:EMNLP} and
\cite{dyer2015}, we use reported results; the other baselines were run
by Bernd Bohnet using version 3.3.0 of the Stanford dependencies and
our predicted POS tags for all datasets to make comparisons as fair as
possible. On the WSJ and Web tasks, our parser outperforms all
dependency parsers in our comparison by a substantial margin. The
Question (QTB) dataset is more sensitive to the smaller beam size we
use in order to train the models in a reasonable time; if we increase
to $B=32$ at inference time only, our perceptron performance goes up
to 92.29\% LAS.

Since many of the baselines could not be directly compared to our
semi-supervised approach, we re-implemented
\newcite{zhang-nivre:2011:ACL} and trained on the tri-training
corpus. Although tri-training did help the baseline on the dev set
(Figure \ref{fig:uptraining_results}), test set performance did not
improve significantly. In contrast, it is quite exciting to see that
after tri-training, even our greedy parser is more accurate than any
of the baseline dependency parsers and competitive with the
BerkeleyParser used to generate the tri-training data. As expected,
tri-training helps most dramatically to increase accuracy on the
Treebank Union setup with diverse domains, yielding 0.4-1.0\% absolute
LAS improvement gains for our most accurate model.

Unfortunately we are not able to compare to several semi-supervised dependency
parsers that achieve some of the highest reported accuracies on the WSJ, in
particular \newcite{suzuki-EtAl:2009:EMNLP}, \newcite{suzuki2011learning} and
\newcite{chen-etAl:2013:EMNLP}.  These parsers use the
\newcite{yamada2003statistical} dependency conversion and the accuracies are
therefore not directly comparable. The highest of these is
\newcite{suzuki2011learning}, with a reported accuracy of 94.22\% UAS. Even
though the UAS is not directly comparable, it is typically similar, and this
suggests that our model is competitive with some of the highest reported
accuries for dependencies on WSJ.

\section{Discussion}
\label{sec:discussion}

In this section, we investigate the contribution of the various components of
our approach through ablation studies and other systematic experiments.  We tune
on Section 24, and use Section 22 for comparisons in order to not pollute the
official test set (Section 23).  We focus on UAS as we found the LAS scores to
be strongly correlated. Unless otherwise specified, we use 200 hidden units in
each layer to be able to run more ablative experiments in a reasonable amount of
time.

\subsection{Impact of Network Structure}
\label{sec:network_structure}

In addition to initialization and hyperparameter tuning, there are several
additional choices about model structure and size a practitioner faces when
implementing a neural network model.  We explore these questions and justify the
particular choices we use in the following. Note that we do not use a beam for
this analysis and therefore do not train the final perceptron layer. This is
done in order to reduce training times and because the trends persist across
settings.

\paragraph{Variance reduction with pre-trained embeddings.} Since the
learning problem is non-convex, different initializations of the
parameters yield different solutions to the learning problem. Thus,
for any given experiment, we ran multiple random restarts for every
setting of our hyperparameters and picked the model that performed
best using the held-out tune set. We found it important to allow the
model to stop training early if tune set accuracy decreased.

\begin{figure}[t]
\centering
\includegraphics[width=0.45\textwidth]{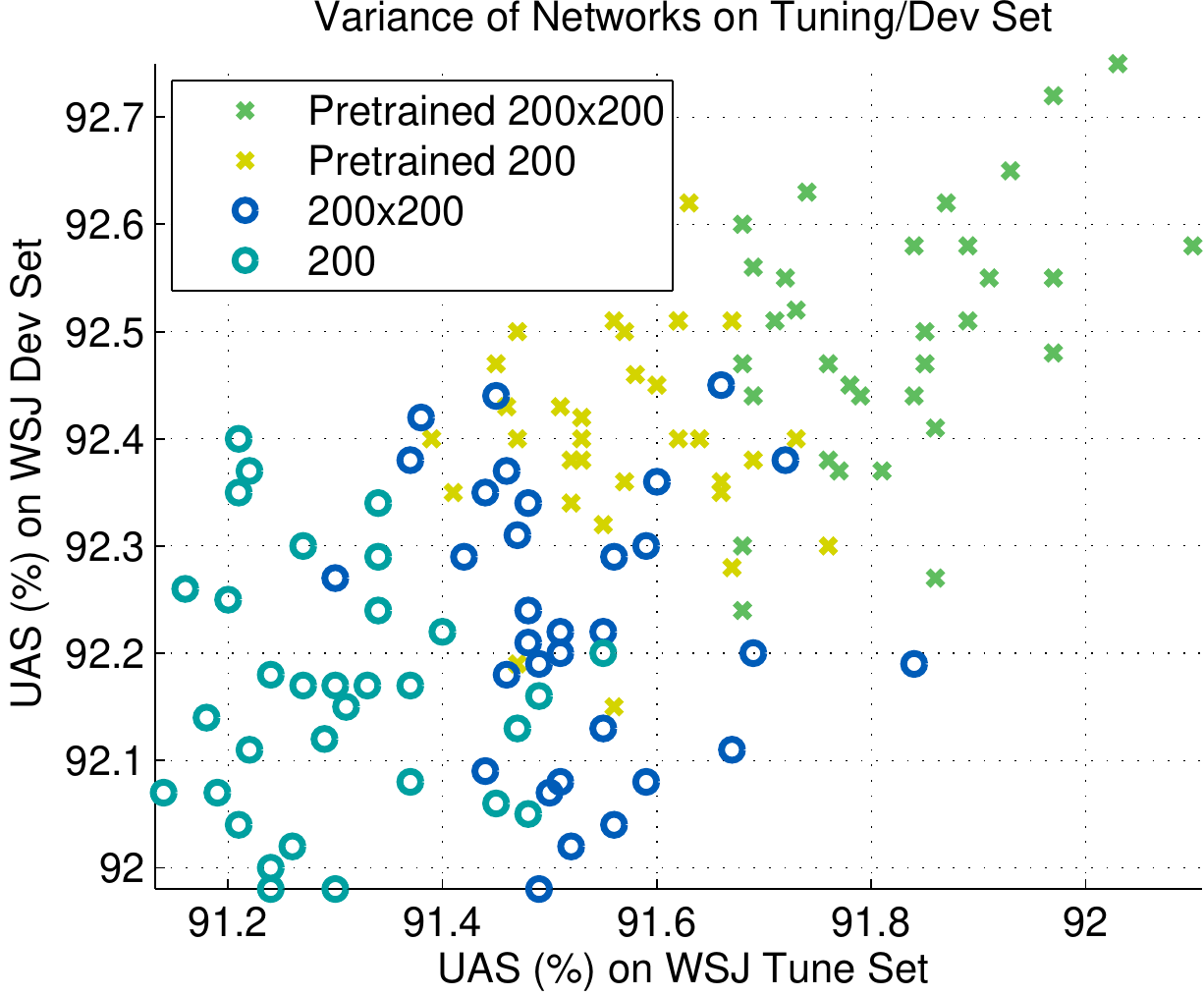}
\caption{\label{fig:variance} Effect of hidden layers and pre-training on
  variance of random restarts. Initialization was either completely random or
  initialized with {\tt word2vec} embeddings (``Pretrained''), and either one or
  two hidden layers of size 200 were used (``200'' vs ``200x200''). Each point
  represents maximization over a small hyperparameter grid with early stopping
  based on WSJ tune set UAS score. $D_{\mwords} = 64$,
  $D_{\mtags},D_{\mlabels} = 16$.}
\end{figure}

We visualize the performance of 32 random restarts with one or two
hidden layers and with and without pretrained word embeddings in
Figure \ref{fig:variance}, and a summary of the figure in Table
\ref{tab:variance}. While adding a second hidden layer
results in a large gain on the tune set, there is no gain on the dev
set if pre-trained embeddings are not used. In fact, while the overall
UAS scores of the tune set and dev set are strongly correlated
($\rho = 0.64$, $p<10^{-10}$), they are {\em not} significantly
correlated if pre-trained embeddings are not used ($\rho = 0.12$,
$p > 0.3$).  This suggests that an additional benefit of pre-trained
embeddings, aside from allowing learning to reach a more accurate
solution, is to push learning towards a solution that generalizes to
more data.

\begin{table}[t]
  \centering
  \scalebox{0.9}{
  \begin{tabular}[h]{cccc}
    \hline
    Pre & Hidden & WSJ \S 24 (Max) & WSJ \S 22 \\ \hline \hline
    Y & $200 \times 200$ & $92.10 \pm 0.11$ & {\bf 92.58} $ \pm 0.12$ \\
    Y & $200$ & $91.76 \pm 0.09$ & $92.30 \pm 0.10$ \\
    N & $200 \times 200$ & $91.84 \pm 0.11$ & $92.19 \pm 0.13$ \\
    N & $200$ & $91.55 \pm 0.10$ & $92.20 \pm 0.12$ \\
    \hline
  \end{tabular}}
\caption{Impact of network architecture on UAS for greedy inference.
  We select the best model from 32 random
  restarts based on the tune set and show the resulting dev set accuracy.
  We also show the standard deviation across the 32 restarts.}
  \label{tab:variance}
\end{table}

\begin{figure*}
\centering
\includegraphics[width=0.45\textwidth]{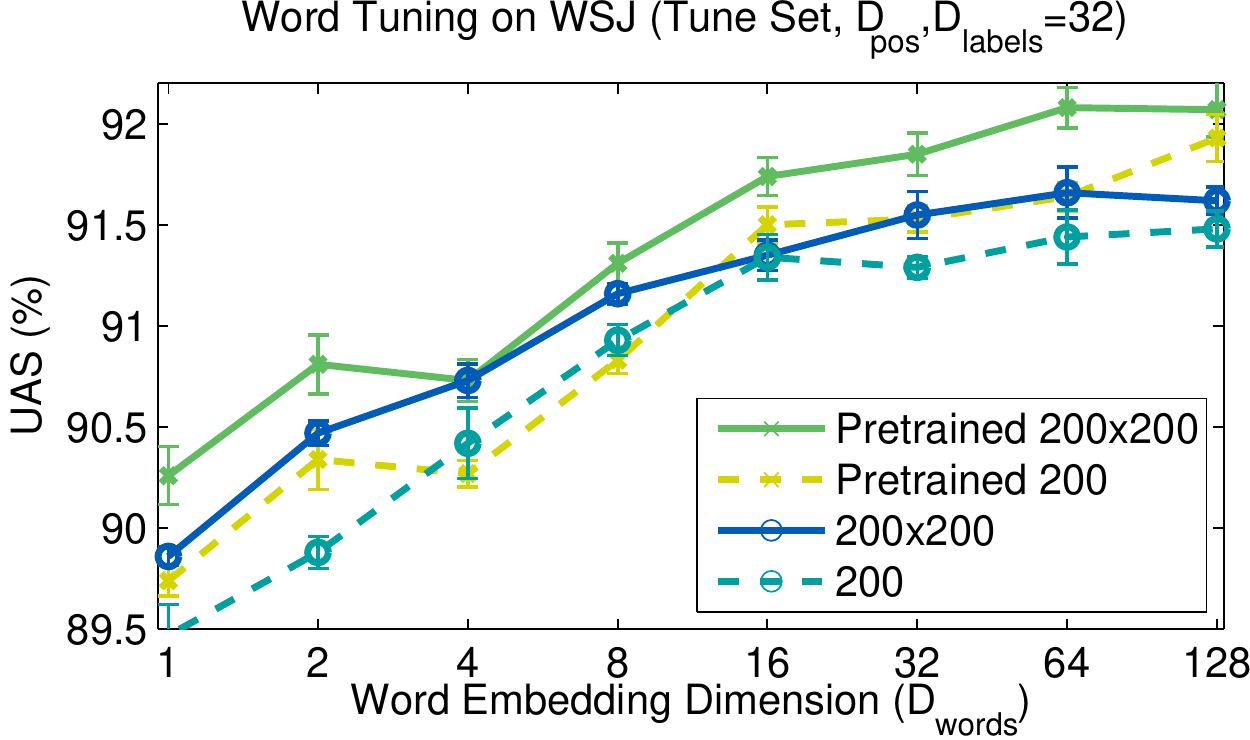}
\includegraphics[width=0.45\textwidth]{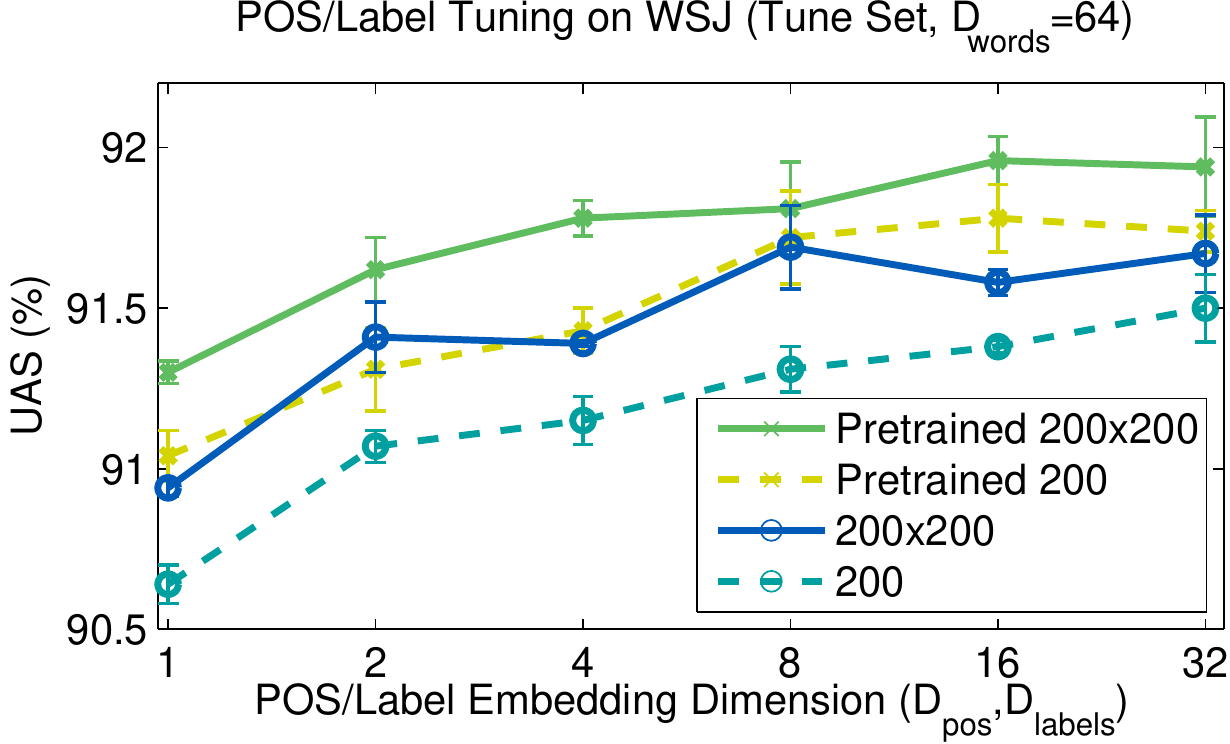}
\vspace{-0.8em}
\caption{\label{fig:embedding_dimensions}
Effect of embedding dimensions on the WSJ tune set.}
\end{figure*}

\paragraph{Diminishing returns with increasing embedding dimensions.}
For these experiments, we fixed one embedding type to a high value and
reduced the dimensionality of all others to very small values. The
results are plotted in Figure \ref{fig:embedding_dimensions},
suggesting larger embeddings do not significantly improve results. We
also ran tri-training on a very compact model with $D_{\mwords} = 8$
and $D_{\mtags} = D_{\mlabels} = 2$ (8$\times$ fewer parameters than
our full model) which resulted in 92.33\% UAS accuracy on the dev set.
This is comparable to the full model without tri-training, suggesting
that more training data can compensate for fewer parameters.

\paragraph{Increasing hidden units yields large gains.} For these
experiments, we fixed the embedding sizes $D_{\mwords} = 64$,
$D_{\mtags} = D_{\mlabels} = 32$ and tried increasing and decreasing
the dimensionality of the hidden layers on a logarthmic
scale. Improvements in accuracy did not appear to saturate even with
increasing the number of hidden units by an order of magnitude, though
the network became too slow to train effectively past $M = 2048$.
These results suggest that there are still gains to be made by
increasing the efficiency of larger networks, even for greedy
shift-reduce parsers.

\begin{table}[t]
  \centering
  \scalebox{0.9}{
  \renewcommand{\arraystretch}{1.0}
  \setlength\tabcolsep{2.2pt}
  \begin{tabular}[h]{lcccccc}
    \hline
    \# Hidden & 64 & 128 & 256 & 512 & 1024 & 2048 \\
    \hline
    \hline
    1 Layer & 91.73 & 92.27 & 92.48 & 92.73 & 92.74 & {\bf 92.83} \\
    2 Layers & 91.89 & 92.40 & 92.71 & 92.70 & 92.96 & {\bf 93.13} \\
    \hline
  \end{tabular}}
\caption{Increasing hidden layer size increases WSJ Dev UAS.
  Shown is the average WSJ Dev UAS  across hyperparameter tuning
  and early stopping with 3 random restarts with a greedy model. }
  \label{tab:beam}
\end{table}

\begin{table}[t]
  \centering
  \scalebox{0.9}{
  \renewcommand{\arraystretch}{1.0}
  \setlength\tabcolsep{2.2pt}
  \begin{tabular}[h]{lcccccc}
    \hline
    Beam & 1 & 2 & 4 & 8 & 16 & 32 \\
    \hline
    \hline
    \multicolumn{7}{l}{\em WSJ Only} \\
    \quad ZN'11      & 90.55 & 91.36 & 92.54 & 92.62 & 92.88 & {\bf 93.09} \\
    \quad Softmax    & 92.74 & 93.07 & 93.16 & {\bf 93.25} & 93.24 & 93.24  \\
    \quad Perceptron & 92.73 & 93.06 & 93.40 & 93.47 & 93.50 & {\bf 93.58} \\
    \multicolumn{7}{l}{\em Tri-training} \\
    \quad ZN'11      & 91.65 & 92.37 & {\bf 93.37} & 93.24 & 93.21 & 93.18 \\
    \quad Softmax    & 93.71 & 93.82 & 93.86 & {\bf 93.87} & 93.87 & 93.87 \\
    \quad Perceptron & 93.69 & 94.00 & 94.23 & {\bf 94.33} & 94.31 & 94.32 \\
    \hline
  \end{tabular}}
  \caption{Beam search always yields significant gains but
    using perceptron training provides even larger benefits, especially for
    the tri-trained neural network model.
    The best result for each model is highlighted in bold.}
  \label{tab:beam}
\end{table}

\begin{table}[t]
  \centering
  \scalebox{0.9}{
  \setlength\tabcolsep{2.2pt}
  \begin{tabular}[h]{ccc}
    \hline
    $\phi(x,c)$ & WSJ Only & Tri-training \\
    \hline
    \hline
    $[\bh_2]$ & 93.16 & 93.93 \\
    $[ P(y) ]$ & 93.26 & 93.80 \\
    $[\bh_1~\bh_2]$ & 93.33 & 93.95 \\
    $[\bh_1~\bh_2~P(y)]$ & {\bf 93.47} & {\bf 94.33} \\
    \hline
  \end{tabular}}
  \caption{Utilizing all intermediate representations improves performance on the WSJ dev set. All results are with $B=8$.}
  \label{tab:hidden}
\end{table}

\subsection{Impact of Structured Perceptron}
\label{sec:perceptron_results}
We now turn our attention to the importance of
structured perceptron training 
as well as the impact of different latent representations.

\paragraph{Bias reduction through structured training.}
To evaluate the impact of structured training, we compare using
the estimates $P(y)$ from the neural network directly for beam search
to using the activations
from all layers as features in the structured perceptron.
Using the probability estimates directly is very similar to
\newcite{ratnaparkhi-parser}, where a maximum-entropy model was used
to model the distribution over possible actions at each parser state,
and beam search was used to search for the highest probability
parse. A known problem with beam search in this setting is the
label-bias problem.
Table \ref{tab:beam} shows the impact of using structured perceptron
training over using the softmax function during beam search as a
function of the beam size used. For reference, our reimplementation of
\newcite{zhang-nivre:2011:ACL} is trained equivalently for each
setting. We also show the impact on beam size when tri-training is
used. Although the beam does marginally improve
accuracy for the softmax model, much greater gains are achieved when
perceptron training is used.

\paragraph{Using all hidden layers crucial for structured perceptron.}
We also investigated the impact of connecting the final perceptron
layer to all prior hidden layers (Table \ref{tab:hidden}). Our results
suggest that all intermediate layers of the network are indeed
discriminative. Nonetheless, aggregating all of their activations proved
to be the most effective representation for the structured
perceptron. This suggests that the representations learned by the
network collectively contain the information required to reduce the
bias of the model, but not when filtered through the softmax
layer. Finally, we also experimented with connecting both hidden
layers to the softmax layer during backpropagation training, but we
found this did not significantly affect the performance of the greedy
model.

\subsection{Impact of Tri-Training}

\begin{figure}[t]
  \centering
  \includegraphics[width=0.45\textwidth]{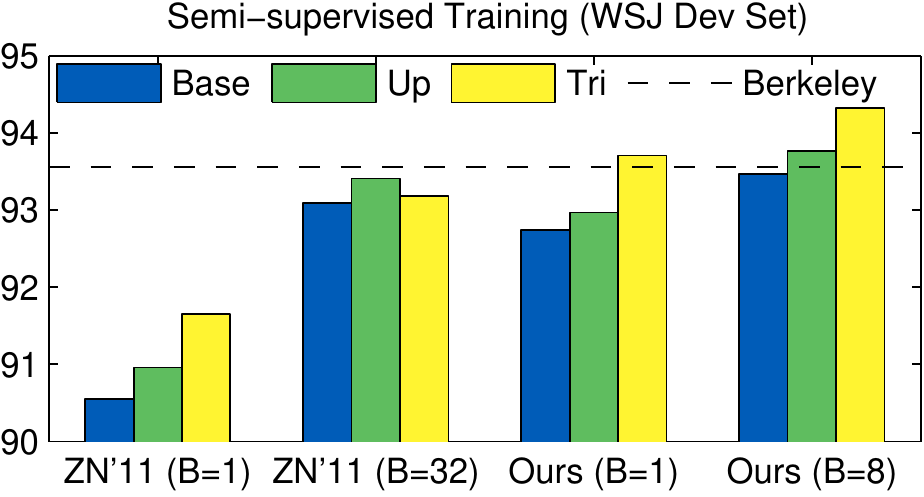}
  \caption{Semi-supervised training with $10^7$ additional tokens,
  showing that tri-training gives significant improvements over up-training
  for our neural net model.
  \label{fig:uptraining_results}}
\end{figure}

To evaluate the impact of the tri-training approach, we compared to
up-training with the BerkelyParser
\cite{petrov-EtAl:2006:ACL} alone. The results are summarized in
Figure \ref{fig:uptraining_results} for the greedy and perceptron
neural net models as well as our reimplementated
\newcite{zhang-nivre:2011:ACL} baseline.

For our neural network model, training on the output of the
BerkeleyParser yields only modest gains, while training on the data
where the two parsers agree produces significantly better
results. This was especially pronounced for the greedy models: after
tri-training, the greedy neural network model surpasses the
BerkeleyParser in accuracy. It is also interesting to note that
up-training improved results far more than tri-training for
the baseline. We speculate that this is due to the a lack of
diversity in the tri-training data for this model, since the same
baseline model was intersected with the BerkeleyParser to generate the
tri-training data.

\subsection{Error Analysis}

Regardless of tri-training, using the structured perceptron improved
error rates on some of the common and difficult labels: {\tt ROOT}, {\tt
  ccomp}, {\tt cc}, {\tt conj}, and {\tt nsubj} all improved by
$>$1\%. We inspected the learned perceptron weights $\mathbf{v}$ for
the softmax probabilities $P(y)$ (see Appendix) and found that the
perceptron reweights the softmax probabilities based on common
confusions; e.g. a strong negative weight for the action RIGHT({\tt
  ccomp}) given the softmax model outputs RIGHT({\tt conj}). Note that
this trend did not hold when $\phi(x,c) = [P(y)]$; without the hidden
layer, the perceptron was not able to reweight the softmax
probabilities to account for the greedy model's biases.


\section{Conclusion}

We presented a new state of the art in dependency parsing: a transition-based
neural network parser trained with the structured perceptron and ASGD. We then
combined this approach with unlabeled data and tri-training to further push
state-of-the-art in semi-supervised dependency parsing. Nonetheless, our ablative
analysis suggests that further gains are possible simply by scaling up our
system to even larger representations. In future work, we will apply our method
to other languages, explore end-to-end training of the system using structured
learning, and scale up the method to larger datasets and network structures.

\section*{Acknowledgements}
We would like to thank Bernd Bohnet for training his parsers and TurboParser
on our setup. This paper benefitted tremendously from discussions with
Ryan McDonald, Greg Coppola, Emily Pitler and Fernando Pereira. Finally,
we are grateful to all members of the Google Parsing Team.

\balance
\bibliographystyle{acl}
\bibliography{acl2015}

\end{document}


\maketitle

\appendix
\section{Full Treebank Union Results}

\paragraph{Corpus details.} The corpora used in the Treebank Union setup are:
\begin{itemize}
\item English News Text Treebank: Penn Treebank Revised (LDC2015T13),
\item OntoNotes version 5 \cite{hovy-EtAl:2006:NAACL},
\item English Web Treebank \cite{petrov-mcdonald:2012:SANCL},
\item Question Treebank \cite{judge-etAl:2006:ACL} (updated and corrected),
\end{itemize}
yielding roughly $\sim90$K training sentences in total.

The full test results including both UAS and LAS are included in Table \ref{tab:treebank_union}.
\begin{table*}[t]
  \centering
  \begin{tabular}{lccccccc}
    \hline
    & & \multicolumn{2}{c}{News} & \multicolumn{2}{c}{Web} & \multicolumn{2}{c}{Questions} \\
    Method & Beam & UAS & LAS & UAS & LAS & UAS & LAS \\
    \hline
      {\em Graph-based} & & & \\
      \quad \newcite{bohnet:2010:COLING} & n/a & 93.29 & 91.38 & 88.22 & 85.22 & 94.01 & 91.49 \\
      \quad \newcite{martins-etAl:2013:ACL} & n/a & 93.10 & 91.13 & 88.23 & 85.04 & 94.21 & 91.54 \\
      \quad \newcite{zhang-mcdonald:2014:ACL} & n/a & 93.32 & 91.48 & 88.65 & 85.59 & 93.37 & 90.69 \\
      \midrule
      {\em Transition-based} & & & \\
      \quad $^\star$\newcite{zhang-nivre:2011:ACL} & 32 & 92.99 & 91.15 & 88.09 & 85.24 & {\bf 94.38} & {\bf 92.46}  \\
      \quad \newcite{bohnet2012best} & 40 & 93.35  & 91.69 & 88.32 & 85.33 & 93.87 & 92.21 \\
      \quad Our Greedy & 1 & 92.92 & 91.21 & 88.32 & 85.41 & 92.79 & 90.61  \\
      \quad Our Perceptron & 16 & {\bf 93.91} & {\bf 92.25} & {\bf 89.29} & {\bf 86.44} & 94.17 & 92.06 \\
      \midrule
      \midrule
      {\em Tri-training} & & & \\
      \quad  $^\star$\newcite{zhang-nivre:2011:ACL} & 32 & 93.22 & 91.46 & 88.40 & 85.51 & 93.74 & 91.36 \\
      \quad  Our Greedy & 1 &  93.48 & 91.82 & 89.18 & 86.37 & 92.60 & 90.58 \\
      \quad  Our Perceptron & 16 & {\bf 94.16 }& {\bf 92.62} & {\bf 89.72} & {\bf 87.00} & {\bf 95.58} & {\bf 93.05} \\
      \bottomrule

    \hline
  \end{tabular}
  \caption{\label{tab:treebank_union}Final Treebank Union test set results. We report LAS only for brevity; see Appendix for full results. For these tri-training results, we sampled sentences to ensure the distribution of sentence lengths matched the distribution in the training set, which we found marginally improved the ZPar tri-training performance.
    For reference, the accuracy of the Berkeley constituency parser (after conversion) is 93.29\% / 91.66\% News, 88.77\% / 85.93\%  Web, and 94.92\% / 93.45\% QTB.}
\end{table*}

\section{Visualization of perceptron weights.}

The perceptron weights $\mathbf{v}$ from the $\phi(x,c) = [\bh_1 \bh_2 P(y)]$ are
visualized in the attached {\tt weights\_full.png}, and for $\phi(,x) = [P(y)]$ in {\tt weights\_Py\_only.png}.

\bibliographystyle{acl}
\bibliography{acl2015}